# Design, Modeling, and Control of Norma:
# a Slider & Pendulum-Driven Spherical Robot

Saeed Moazami[1], Srinivas Palanki[2], and Hassan Zargarzadeh[1]

ABSTRACT

   This paper discusses the design, modeling, and control of Norma, a novel 2-DOF mobile spherical robot (SR). The propelling mechanism of this robot consists of two actuators: a slider, and a rotational pendulum located on the SR's diagonal shaft. The slider can translate along the shaft, shifting the robot's center of gravity towards the robot's sides, and the pendulum rotates around the shaft to propel the SR to roll forward and backward. These two actuators enable the SR to perform rolling and turning maneuvers simultaneously. The advantage of the proposed design lies in its agility and accuracy of the mathematical model. The Euler-Lagrange approach is utilized to derive the dynamics of the proposed structure. Next, a path tracking control scheme is introduced for a smooth trajectory. Finally, simulations are carried out to verify the accuracy of the mathematical model and the effectiveness of the controller.

Keywords:
Spherical Robot, Euler-Lagrange Method, Mathematical Modeling, Dynamics, Mobile Robot.

## 1. INTRODUCTION

   Spherical robots (SRs) are a class of mobile robots that are generally identified by their ball-shaped shell and internal driving components that provide torques required for their rolling motion [1-4]. Despite their limitations, due to their ball-shaped exterior, SRs inherit multiple advantages where skidding, tipping over, falling, or friction with the surface makes other types of mobile robots vulnerable or inefficient. Furthermore, incorporating particular considerations into the design of SRs, all internal components of the robot can be protected from collisions and environmental states such as moisture, radiation, dust, etc. These unique features make SRs a suitable candidate for various applications such as surveillance [5], environmental monitoring [6], exploration operations in unknown environments [7], agriculture [8], rehabilitation [9], and hobbies [10]. Nonetheless, compared to other types of mobile robots, e.g., wheeled and legged robots, SRs are by far underemphasized by researchers in the field of mobile robotics. One reason can be the complicated dynamics of SRs that has remained a hurdle to comprehend their behavior and maneuverability features.

   There are a number of SR classifications available in the literature. In [1], SRs are classified based on their mechanical configurations, e.g., wheeled [11], pendulum driven [12-16], gimbal mechanism [17], single ball [1], mass movement [18], and a set of designs that use flywheels [19, 20]. SRs can also be classified based on their kinematics behavior [4] into Continuous Rolling Spherical Robots (CR-SR), and Rolling and Steering Spherical Robots (RS-SR). Alternatively, in [2], SRs are classified based on their mechanical driving principles as 1) conservation of angular momentum (COAM), in which SRs utilize flywheels to provide propelling torques, 2) outer shell transformation (OST), that the robots deform different parts of their outer shell in order to propel themselves, and 3) barycenter offset (BCO), where SRs shift their center of gravity to provide gravitational torque. It can be noted that a majority of SRs belong to BCO SRs and one of the most popular structures in this class is pendulum-driven spherical robot (PDSR).

[1] Phillip M. Drayer Department of Electrical Engineering at Lamar University, Beaumont, TX, USA. (e-mails: smoazami@lamar.edu, h.zargar@lamar.edu).
[2] Dan F. Smith Department of Chemical Engineering at Lamar University, Beaumont, TX, USA. (e-mail: spalanki@lamar.edu).

PDSRs are usually modeled as a spherical shell with a diagonal shaft and a pendulum mounted on the shaft, rotating about the shaft's axis. In the literature, several mechanisms of PDSRs have been proposed that utilize either a single 2-DOF pendulum [21-24] or double-pendulum structure [25-27] as the source of gravitational torque. In [28], a mathematical model of a PDSR is derived using a decoupled dynamics approach. From a different view, the model of a PDSR is studied in [29, 30], where the desired path is assumed to be a straight line with a constant slope and a 2D curved path with a variable-slope respectively. Likewise, in [31], the mathematical model is derived for an SR using the Lagrangian reduction theory defined on symmetry groups.

Taking a step forward, with the aim to introduce a novel SR model, this paper investigates the modeling and control of Norma, an SR with a 2-DOF propelling mechanism which is comprised of a linear and a rotational actuator located on a diagonal shaft fixed to a solid spherical shell. The rotational actuator provides gravitational torque for rolling action of the SR along the longitudinal direction by applying torque through a pendulum to the shaft. The linear actuator translates a sliding weight on the shaft that shifts the center of gravity of the robot in the transverse direction causing the robot to tilt and consequently providing the turning maneuverability for the robot. To derive the mathematical model of Norma, first, a conceptual model is presented. Then, the Euler-Lagrange method is utilized to derive the dynamics of the proposed SR.

The path tracking control of SRs has been a challenge due to its underactuated features and complicated dynamics. In this paper, this problem is addressed by a control scheme that decouples the SR's kinematics from its dynamics. To that end, first, a set of PID controllers are designed that stabilize the SR's rolling angular velocity, $\dot{\theta}$, and its slider displacement, $\delta$, around and along the transverse axis, respectively. Then, a path planning algorithm is devised that governs $\dot{\theta}$ and $\delta$ based on the path tracking error and the robot's physical constraints. Finally, simulations are carried out to evaluate the SR model's accuracy and the controller's tracking performance. The conducted analyses show that the controller can control the proposed SR model successfully.

The rest of the paper is organized into the following sections: the description of the proposed model is given in the second section; in the third section, kinematics and dynamics models are developed; in section four, the control scheme is presented. The simulation results are represented in section five, followed by the conclusions in the last section.

## 2. MODEL DESCRIPTION

Fig. 1 illustrates the schematic diagram of the proposed SR. The robot consists of a ball-shaped rigid shell, a diagonal shaft, a linear, and a rotational actuator. The shaft is along the transverse axis of the sphere, and it is fixed to the shell. In order to provide the rolling motion, the pendulum can apply torque $\mathcal{T}$ to the shaft around the transverse axis. This torque causes the sphere to roll forward and backward along the longitudinal axis which is horizontal and perpendicular to the transverse axis. The slider translates along the shaft's axis using force $\mathcal{F}$. Translation of the slider shifts the center of mass of the spherical robot in the transverse direction, resulting in the shaft and the sphere to tilt about the longitudinal axis. So, the proposed SR has the ability of both rolling towards the longitudinal axis and tilting about the same axis, enabling it to have turning motion.

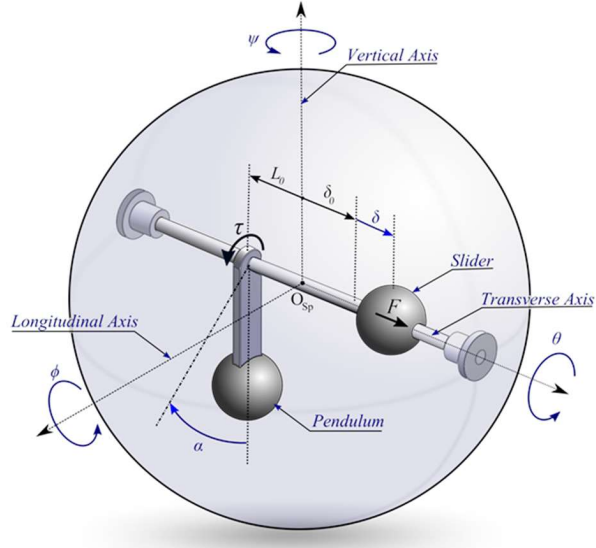

Fig. 1. Schematic diagram of Norma, the proposed SR.

*Assumptions*: The following assumptions are made for deriving the kinematics and dynamics of the SR:
1) *The sphere rolls over a perfectly flat horizontal plane surface without slipping.*
2) *The slider and pendulum bob are point masses, and the pendulum rod is massless.*

When there is no force or torque applied, the SR is designed to have a resting static equilibrium point (RSEP) where the pendulum rod is perpendicular to the ground, and the transverse axis is horizontal. At this point, the plumb line from the sphere center, denoted as $O_{Sp}$, passes through the robot's center of gravity. Considering $m_P$, $L_0$, $m_{Sl}$, $\delta_0$ respectively as the pendulum's mass, the distance of the pendulum's point of suspension from $O_{Sp}$, slider mass, and the slider's distance from $O_{Sp}$ at RSEP, we can write:

$$m_p L_0 = m_{Sl} \delta_o . \tag{1}$$

The next step is to define the SR's generalized coordinates. As shown in Fig. 1, $\theta$ and $\phi$ are the rotation angles of the sphere about the transverse and longitudinal axes, respectively. In fact, $\phi$ is the tilting angle of the sphere. Propelled by torque $\mathcal{T}$, $\alpha$ is the rotation of the pendulum about the shaft which is in the opposite direction to $\theta$. Also, propelled by force $\mathcal{F}$, $\delta$ is the displacement of the slider along the shaft axis with respect to $\delta_0$ as the origin.

In the sequel, reference frames are denoted as $\{\mathcal{F}\}$, where $\mathcal{F}$ is the set of frame axes. The reference frame of any vector quantity is shown in its left superscript e.g., $^{\mathcal{F}}A$ represents $A$ in $\{\mathcal{F}\}$. Rotation matrices are denoted as $\mathcal{R}_{\mathcal{F}_B \mathcal{F}_A}$ which transforms vectors from $\{\mathcal{F}_A\}$ to $\{\mathcal{F}_B\}$.

Fig. 2 illustrates the reference frames that are used to derive the equations of motion of the spherical robot. $\{\mathcal{F}_W\}$ with axes $\{X_W^-, Y_W^-, Z_W^-\}$ is the world coordinate frame fixed to the ground at its origin $O_W$. The position coordinates of the SR in $\{\mathcal{F}_W\}$ are denoted as:

$$^{\mathcal{F}_W} r_{Sp} = {}^{\mathcal{F}_W}\left[X_{Sp}(t), Y_{Sp}(t), 0\right]^T . \tag{2}$$

$\{\mathcal{F}_l\}$ is the local reference frame with axes $\{x_l^-, y_l^-, z_l^-\}$, and its origin $O_l$ located at $O_{Sp}$. $x_l^-$ and $y_l^-$ are along the SR's longitudinal and lateral axes. $z_l^-$ is upward and perpendicular to the ground. $\{\mathcal{F}_S\}$ with axes

$\{x_S^-, y_S^-, z_S^-\}$ is the shaft reference frame whose origin is at $O_S$. $x_S^-$ and $y_S^-$ are along the SR's longitudinal and transverse axes. $z_S^-$ is mutually perpendicular to $x_S^-$ and $y_S^-$ by following the right-hand rule.

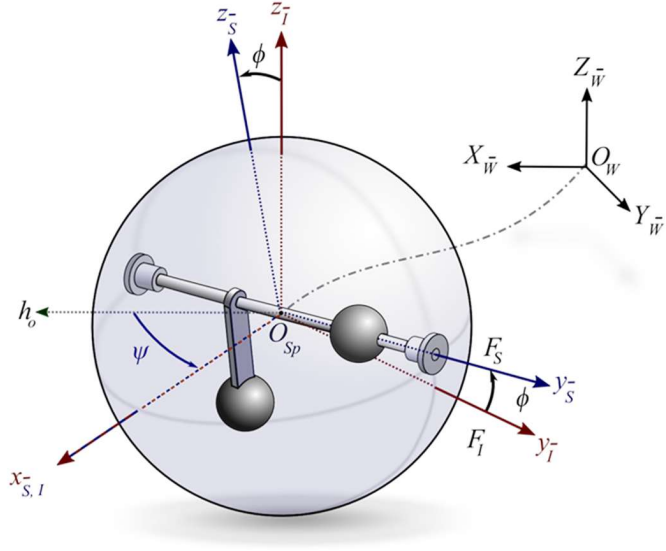

Fig. 2. The reference frames used in the spherical robot kinematics model.

As depicted in Fig. 2, $x_I^-$ and $y_I^-$ are always horizontal, i.e. parallel to $X_W Y_W$ plane and rotated about $z_I^-$ axis by the angle of $\psi$ that is the SR's turning angle about its vertical axis. In other words, the angle between $X_W^-$ and $x_I^-$ is $\psi$. Accordingly, $\{\hat{I}, \hat{J}, \hat{K}\}$, $\{\hat{i}, \hat{j}, \hat{k}\}$, and $\{\hat{i}_S, \hat{j}_S, \hat{k}_S\}$ represent unit vectors of $\{\mathcal{F}_W\}, \{\mathcal{F}_I\}$, and $\{\mathcal{F}_S\}$ respectively. The next section is dedicated to the derivation of the kinematics and dynamics of Norma.

## 3. KINEMATICS AND DYNAMICS MODELING OF NORMA

In this section, first, the kinematics equations of the SR's turning maneuver are developed. Then, using the Euler-Lagrange method, dynamics equations of motion of the robot are derived.

### A. Kinematics and Modeling of Turning Action

Norma's turning action is modeled in Fig. 3. By considering the SR at tilting angle of $\phi$, using an imaginary cone with an apex angle of $2\phi$ that is purely rolling over the ground. The cone's instantaneous axis of rotation is the imaginary contact line of the cone with the ground, passing through the vertex of the cone, $V$, and the SR point of contact with the ground $C_P$. Based on the assumption (1) and according to the equal-arc-length rule for rolling without slipping [32] for the cone's base circle, we have:

$$r_c \dot{\theta} = \rho \dot{\psi}, \tag{3}$$

where $r_c$ is cone's base circle's radius, and $\rho$ is the instantaneous radius of curvature of the turning motion that is the distance between the vertex of the cone and the contact point. Considering $R$ as sphere's radius we have:

$$r_c = R C_\phi, \tag{4}$$

$$\rho = -R C_\phi / S_\phi. \tag{5}$$

Where $C_\bullet$ and $S_\bullet$ represent $\cos(\bullet)$ and $\sin(\bullet)$ respectively. Substituting (4) and (5) in (3) results in:

$$\dot{\psi} = -\dot{\theta} S_\phi. \tag{6}$$

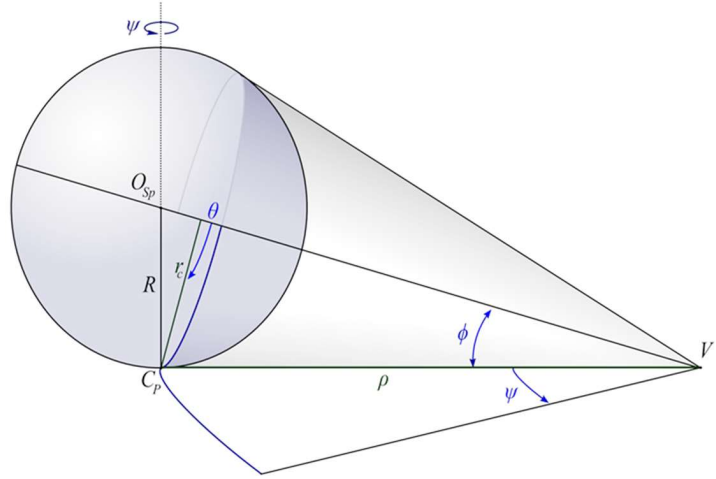

Fig. 3. Rolling cone model of the spherical robot's turning maneuver.

Now, the position vector of the $O_{Sp}$ in $\{\mathcal{F}_I\}$ with respect to $C_P$ can be written as:

$$^{\mathcal{F}_I}\overline{C_P O_{Sp}} = R\hat{k}, \tag{7}$$

Also, according to the definition of $\theta$ and $\phi$, and from (6), the sphere's angular velocity is written as:

$$^{\mathcal{F}_I}\Omega_{Sp} = \dot{\phi}\hat{i} + \dot{\theta}C_\phi \hat{j} - \dot{\theta}S_\phi \hat{k}. \tag{8}$$

Then, the linear velocity vector of $O_{Sp}$ can be calculated in $\{\mathcal{F}_I\}$ as the following:

$$^{\mathcal{F}_I}V_{Sp} = {^{\mathcal{F}_I}\Omega_{Sp}} \times {^{\mathcal{F}_I}r_{C_P O}} = R\dot{\theta}C_\phi \hat{i} - R\dot{\phi}\hat{j}. \tag{9}$$

In $^{\mathcal{F}_I}V_{Sp}$, the sphere velocity in $z_I^-$ direction is $^{\mathcal{F}_I}w_{Sp} = 0$, so it can be denoted as:

$$^{\mathcal{F}_I}V_{Sp} = {^{\mathcal{F}_I}[u_{Sp}, v_{Sp}, 0]^T}. \tag{10}$$

As shown in Fig. 2, transformation matrix from $\{\mathcal{F}_I\}$ to $\{\mathcal{F}_W\}$ is a rotation matrix about $z_I^-$ and angle of $\psi$ as follows:

$$\mathcal{R}_{\mathcal{F}_W \mathcal{F}_I} = \begin{bmatrix} C_\psi & -S_\psi & 0 \\ S_\psi & C_\psi & 0 \\ 0 & 0 & 1 \end{bmatrix}. \tag{11}$$

Therefore, using (9) - (11), one can write the velocity of $O_{Sp}$ in $\{\mathcal{F}_W\}$ as $^{\mathcal{F}_W}V_{Sp} = \mathcal{R}_{\mathcal{F}_W \mathcal{F}_I} {^{\mathcal{F}_I}V_{Sp}}$ that results in:

$$\begin{aligned} ^{\mathcal{F}_W}\dot{X}_{Sp} &= R\dot{\theta}C_\phi C_\psi + R\dot{\phi}S_\psi, \\ ^{\mathcal{F}_W}\dot{Y}_{Sp} &= R\dot{\theta}C_\phi S_\psi - R\dot{\phi}C_\psi. \end{aligned} \tag{12}$$

So far, the angular velocity of SR about its vertical axis is calculated, and regardless of the dynamic states of the elements in the SR, the velocity of the robot is introduced in the local and global reference frames.

*B. Motion Dynamics of SR*

Using the Euler-Lagrange method, the motion dynamics of the robot is represented as:

$$\frac{d}{dt}(\frac{\partial \mathcal{L}}{\partial \dot{q}_i}) - \frac{\partial \mathcal{L}}{\partial q_i} = \mathcal{U}_i, \text{ for } i = 1, 2, 3, 4, \tag{13}$$

where $q = [\theta, \alpha, \phi, \delta]^T$ is the set of independent generalized coordinates that describe the robot's configuration and $\mathcal{U} = [\mathcal{U}_\theta, \mathcal{U}_\alpha, \mathcal{U}_\phi, \mathcal{U}_\delta]^T$ are external torques and forces that are applied to the corresponding generalized coordinates. $\mathcal{L}$ is the Lagrangian function that can be expressed as:

$$\mathcal{L}(q,\dot{q}) = E^k(q,\dot{q}) - E^p(q), \qquad (14)$$

where $E^k$ and $E^p$ denote the total kinetic and potential energies of the robot, respectively.

To calculate $E^p$, since the elevation of the sphere does not vary in $\{\mathcal{F}_I\}$, only the slider and the pendulum of the SR contribute to $E^p$, whose elevations are derived in the following. In Fig. 1, the pendulum's rotation around the shaft is denoted by $\alpha$ in the opposite direction to the sphere's rolling rotation, $\theta$. Also, consider the initial direction of the pendulum to be in RSEP. In that case, the difference between the two angles $\alpha - \theta$ measures the angle between the pendulum and its initial direction. So, the position vector $^{\mathcal{F}_S}r_P$ of the pendulum bob in $\{\mathcal{F}_S\}$ can be written as follows:

$$^{\mathcal{F}_S}r_P = LS_{\alpha\theta}\hat{i}_S - L_0\hat{j}_S - LC_{\alpha\theta}\hat{k}_S, \qquad (15)$$

where $L$ is the pendulum's length, $S_{\alpha\theta}$ and $C_{\alpha\theta}$ represent $\sin(\alpha-\theta)$ and $\cos(\alpha-\theta)$, respectively. Furthermore, as shown in Fig. 2, $\{\mathcal{F}_S\}$ is resulted from the rotation of $\{\mathcal{F}_I\}$ by angle of $\phi$ about $x_I^-$. Thus, defining $\mathcal{R}_{\mathcal{F}_I\mathcal{F}_S}$ as:

$$\mathcal{R}_{\mathcal{F}_I\mathcal{F}_S} = \begin{bmatrix} 1 & 0 & 0 \\ 0 & C_\phi & S_\phi \\ 0 & -S_\phi & C_\phi \end{bmatrix}, \qquad (16)$$

from (15) and utilizing (16), the pendulum's position in $\{\mathcal{F}_I\}$ can be written as follows:

$$^{\mathcal{F}_I}r_P = \mathcal{R}_{\mathcal{F}_I\mathcal{F}_S}\,^{\mathcal{F}_S}r_P = LS_{\alpha\theta}\hat{i} + \left(-L_0 C_\phi + LC_{\alpha\theta}S_\phi\right)\hat{j} - \left(LC_{\alpha\theta}C_\phi + L_0 S_\phi\right)\hat{k}. \qquad (17)$$

Based on the definition of $\delta$ and $\delta_0$, the position of the slider in $\{\mathcal{F}_S\}$ is:

$$^{\mathcal{F}_S}r_{Sl} = (\delta+\delta_0)\hat{j}_S, \qquad (18)$$

that can be represented in $\{\mathcal{F}_I\}$ as:

$$^{\mathcal{F}_I}r_{Sl} = \mathcal{R}_{\mathcal{F}_I\mathcal{F}_S}\,^{\mathcal{F}_S}r_{Sl} = (\delta+\delta_0)C_\phi\hat{j} + (\delta+\delta_0)S_\phi\hat{k}. \qquad (19)$$

Now, using the pendulum bob and slider positions in (17) and (19) the potential energy of the robot can be written as the following:

$$E^p = m_{Sl}g(\delta+\delta_0)S_\phi - m_P g\left(LC_{\alpha\theta}C_\phi + L_0 S_\phi\right). \qquad (20)$$

In the calculation of $E^k$, the kinetic energy of all the SR components, namely the sphere, the pendulum, and the slider, denoted respectively as $E^k_{Sp}$, $E^k_P$, and $E^k_{Sl}$, contribute:

$$E^k = E^k_{Sp} + E^k_P + E^k_{Sl}, \text{ with}$$
$$E^k_{Sp} = \tfrac{1}{2}\left(m_{Sp}\|{}^{\mathcal{F}_I}V_{Sp}\|^2 + {}^{\mathcal{F}_I}\Omega^T_{Sp}I_{Sp}\,{}^{\mathcal{F}_I}\Omega_{Sp}\right),\ E^k_P = \tfrac{1}{2}\left(m_P\|{}^{\mathcal{F}_I}V_P\|^2 + {}^{\mathcal{F}_I}\Omega^T_P I_P\,{}^{\mathcal{F}_I}\Omega_P\right), \text{ and} \qquad (21)$$
$$E^k_{Sl} = \tfrac{1}{2}\left(m_{Sl}\|{}^{\mathcal{F}_I}V_{Sl}\|^2 + {}^{\mathcal{F}_I}\Omega^T_{Sl}I_{Sl}\,{}^{\mathcal{F}_I}\Omega_{Sl}\right),$$

where $^{\mathcal{F}_I}V$s and $^{\mathcal{F}_I}\Omega$s are linear velocity and angular velocity vectors of each component. In (21), subscripts $Sp$, $P$, and $Sl$ stand for the sphere, pendulum, and the slider respectively. Further, since the sphere shell and the shaft are fixed to each other, they are considered as a single part, $Sp$, with the total mass of $m_{Sp}$. $I_{Sp}$, $I_P$, and $I_{Sl}$ are moments of inertia matrices of the sphere (with the shaft attached), pendulum, and the slider with respect to $\{\mathcal{F}_I\}$. In the following, linear and angular velocities of all the Norma components are derived in order to calculate $E^k$ in (21).

For the sphere, $^{\mathcal{F}_I}\Omega_{Sp}$ and $^{\mathcal{F}_I}V_{Sp}$ are presented in (8)-(9). For the pendulum, $^{\mathcal{F}_I}V_P$ can be written as:

$$^{\mathcal{F}_I}V_P = V_P^{rel} + {}^{\mathcal{F}_I}V_{O_S} + {}^{\mathcal{F}_I}\Omega_{\mathcal{F}_S} \times {}^{\mathcal{F}_I}r_P, \tag{22}$$

where $V_P^{rel}$ is the relative velocity of the pendulum in $\{\mathcal{F}_I\}$ assuming that $\{\mathcal{F}_S\}$ is stationary. $^{\mathcal{F}_I}V_{O_S}$ is the relative linear velocity of the origin of $\{\mathcal{F}_S\}$ in $\{\mathcal{F}_I\}$, and $^{\mathcal{F}_I}\Omega_{\mathcal{F}_S}$ is the angular velocity of $\{\mathcal{F}_S\}$ in $\{\mathcal{F}_I\}$. $^{\mathcal{F}_I}r_P$ is the position vector of the pendulum that is presented in (15). Calculation steps to derive the terms of the equation (22) are presented in the following. The angular velocity of the pendulum in $\{\mathcal{F}_S\}$ is written as follows:

$$^{\mathcal{F}_S}\Omega_P = (\dot{\theta} - \dot{\alpha})\hat{j}_S. \tag{23}$$

Also, $^{\mathcal{F}_S}V$ can be calculated as:

$$^{\mathcal{F}_S}V_P = {}^{\mathcal{F}_S}\Omega_P \times {}^{\mathcal{F}_S}r_P = L(\dot{\alpha} - \dot{\theta})C_{\alpha\theta}\hat{i}_S + L(\dot{\alpha} - \dot{\theta})S_{\alpha\theta}\hat{k}_S. \tag{24}$$

Then, using (16) and (24), we can write:

$$V_P^{rel} = \mathcal{R}_{\mathcal{F}_I\mathcal{F}_S} {}^{\mathcal{F}_S}V_P = L(\dot{\alpha} - \dot{\theta})C_{\alpha\theta}\hat{i} - L(\dot{\alpha} - \dot{\theta})S_{\alpha\theta}S_\phi\hat{j} + L(\dot{\alpha} - \dot{\theta})S_{\alpha\theta}C_\phi\hat{k}. \tag{25}$$

Knowing $^{\mathcal{F}_I}\Omega_{\mathcal{F}_S} = \dot{\phi}\hat{i}$ and using (17) the term $^{\mathcal{F}_I}\Omega_{\mathcal{F}_S} \times {}^{\mathcal{F}_I}r_P$ in (22) takes the following form:

$$^{\mathcal{F}_I}\Omega_{\mathcal{F}_S} \times {}^{\mathcal{F}_I}r_P = (L_0 S_\phi + LC_{\alpha\theta}C_\phi\dot{\phi})\hat{j} + (LC_{\alpha\theta}S_\phi\dot{\phi} - L_0 C_\phi)\hat{k}. \tag{26}$$

Since $O_{Sp}$ and $O_S$ coincide instantaneously, we have $^{\mathcal{F}_I}V_{O_S} = {}^{\mathcal{F}_I}V_{Sp}$ and from (9) we can write:

$$^{\mathcal{F}_I}V_{O_S} = R\dot{\theta}C_\phi\hat{i} - R\dot{\phi}\hat{j}. \tag{27}$$

Now, combining (25), (26), and (27), (22) can be rewritten as:

$$^{\mathcal{F}_I}V_P = \left(R\dot{\theta}C_\phi + L(\dot{\alpha} - \dot{\theta})C_{\alpha\theta}\right)\hat{i} + \left(-L(\dot{\alpha} - \dot{\theta})S_{\alpha\theta}S_\phi + (L_0 S_\phi + LC_{\alpha\theta}C_\phi - R)\dot{\phi}\right)\hat{j}$$
$$+ \left(L(\dot{\alpha} - \dot{\theta})S_{\alpha\theta}C_\phi + (-L_0 C_\phi + LC_{\alpha\theta}S_\phi\dot{\phi})\right)\hat{k}. \tag{28}$$

Moving on, the angular velocity of the pendulum, by utilizing (16) and (23), can be calculated as follows:

$$^{\mathcal{F}_I}\Omega_P = \mathcal{R}_{\mathcal{F}_I\mathcal{F}_S} {}^{\mathcal{F}_S}\Omega_P = \dot{\phi}\hat{i} + (\dot{\theta} - \dot{\alpha})C_\phi\hat{j} - (\dot{\theta} - \dot{\alpha})S_\phi\hat{k}. \tag{29}$$

Similar to the pendulum, the slider's linear velocity is written as:

$$^{\mathcal{F}_I}V_{Sl} = V_{Sl}^{rel} + {}^{\mathcal{F}_I}V_{O_S} + {}^{\mathcal{F}_I}\Omega_{\mathcal{F}_S} \times {}^{\mathcal{F}_I}r_{Sl}, \tag{30}$$

where $V_{Sl}^{rel}$ is the relative velocity of the slider in $\{\mathcal{F}_I\}$, assuming that $\{\mathcal{F}_S\}$ is stationary. Since the motion of the slider is always along the shaft axis, $y_S^-$, the velocity of the slider in $\{\mathcal{F}_S\}$ is:

$$^{\mathcal{F}_S}V_{Sl} = \dot{\delta}\hat{j}_S. \tag{31}$$

Then, from (16) and (31), $V_{Sl}^{rel}$ becomes:

$$V_{Sl}^{rel} = \mathcal{R}_{\mathcal{F}_I\mathcal{F}_S} {}^{\mathcal{F}_S}V_{Sl} = \dot{\delta}C_\phi\hat{j} + \dot{\delta}S_\phi\hat{k}. \tag{32}$$

Also, in , knowing that $^{\mathcal{F}_I}\Omega_{\mathcal{F}_S} = \dot{\phi}\hat{i}$, from (19), the last term of (30) yields to:

$$^{\mathcal{F}_I}\Omega_{\mathcal{F}_S} \times {}^{\mathcal{F}_I}r_{Sl} = -(\delta + \delta_0)S_\phi\dot{\phi}\hat{j} + (\delta + \delta_0)C_\phi\dot{\phi}\hat{k}. \tag{33}$$

Thus, by plugging (27), (32), and (33) into (30), $^{\mathcal{F}_I}V_{Sl}$ can be rewritten as:

$$^{\mathcal{F}_I}V_{Sl} = \left(R\dot{\theta}C_\phi\right)\hat{i} + \left(\dot{\delta}C_\phi + (-R - (\delta + \delta_0)S_\phi)\dot{\phi}\right)\hat{j} + \left(\dot{\delta}S_\phi + (\delta + \delta_0)C_\phi\dot{\phi}\right)\hat{k}. \tag{34}$$

Finally, to derive $^{\mathcal{F}_I}\Omega_{Sl}$, we have:

$$^{\mathcal{F}_S}\Omega_{Sl} = \dot{\phi}\hat{i}_S + \dot{\theta}\hat{j}_S, \tag{35}$$

which can be transferred to $\{\mathcal{F}_I\}$ using (16) as follows:

$$^{\mathcal{F}_I}\Omega_{Sl} = \dot{\phi}\hat{i} + \dot{\theta}C_\phi\hat{j} - \dot{\theta}S_\phi\hat{k}. \tag{36}$$

The next step is to derive the kinetic energy for Norma's components. To from the kinetic energy equation of the sphere, from (8) and (9) we have:

$$E_{Sp}^k = \tfrac{1}{2}m_{Sp}\left((R\dot{\theta}C_\phi)^2 + (-R\dot{\phi})^2\right) + \tfrac{1}{2}I_{Sp}^x\dot{\phi}^2 + \tfrac{1}{2}I_{Sp}^y\left(\dot{\theta}C_\phi\right)^2 + \tfrac{1}{2}I_{Sp}^z\left(-\dot{\theta}S_\phi\right)^2, \qquad (37)$$

where $I_{Sp}^x$, $I_{Sp}^y$, and $I_{Sp}^z$ are the diagonal elements of $I_{Sp}$ that is the sum of the moments of inertia of the shell and the shaft denoted as $I_{Shl}$ and $I_{Shf}$ respectively. Due to the spherical shape of the shell, $I_{Shl}$ is:

$$I_{Shl} = diag[I_{Shl}^x, I_{Shl}^y, I_{Shl}^z], \text{ with } I_{Shl}^x = I_{Shl}^y = I_{Shl}^z = \tfrac{2}{3}m_{Shl}R^2, \qquad (38)$$

where, $m_{Shl}$, $I_{Shl}^y$, and $I_{Shl}^z$ are moments of inertia of the shell about $\{\mathcal{F}_l\}$ axes, and $m_{Shl}$ is the mass of the shell. Considering the shaft as an oblique rod with a length of $2R$, rotated with the angle of $\phi$ about $x_l^-$, $I_{Shf}$ can be written as:

$$I_{Shf} = diag[I_{Shf}^x, I_{Shf}^y, I_{Shf}^z], \text{ with } I_{Shf}^x = \tfrac{1}{3}m_{Shl}R^2, \; I_{Shf}^y = I_{Shf}^x S_\phi^2, \; I_{Shf}^z = I_{Shf}^x C_\phi^2, \qquad (39)$$

where, $I_{Shf}^x$, $I_{Shf}^y$, and $I_{Shf}^z$ are the shaft's moment of inertia about $\{\mathcal{F}_l\}$ axes. Now, from (38) and (39), the moment of inertia of the sphere can be calculated as follows:

$$I_{Sp}^x = I_{Shl} + I_{Shf}^x, \; I_{Sp}^y = I_{Shl} + I_{Shf}^x S_\phi^2, \; I_{Sp}^y = I_{Shl} + I_{Shf}^x C_\phi^2, \qquad (40)$$

To derive $E_P^k$, using (23) and (24) we have:

$$E_P^k = \tfrac{1}{2}m_P\left(\left(R\dot{\theta}C_\phi + L(\dot{\alpha}-\dot{\theta})C_{\alpha\theta}\right)^2 + \left((L_0 S_\phi + LC_{\alpha\theta}C_\phi - R)\dot{\phi} - L(\dot{\alpha}-\dot{\theta})S_{\alpha\theta}S_\phi\right)^2 \right.$$
$$\left. + \left(L(\dot{\alpha}-\dot{\theta})S_{\alpha\theta}C_\phi + (-L_0 C_\phi + LC_{\alpha\theta}S_\phi)\dot{\phi}\right)^2\right) + \tfrac{1}{2}I_P^x\dot{\phi}^2 + \tfrac{1}{2}I_P^y\left((\dot{\theta}-\dot{\alpha})C_\phi\right)^2 + \tfrac{1}{2}I_P^z\left((\dot{\alpha}-\dot{\theta})S_\phi\right)^2, \qquad (41)$$

in which, $I_P^x$, $I_P^y$, and $I_P^z$ are pendulum's moment of inertia about $\{\mathcal{F}_l\}$ axes. Since the pendulum bob is considered to be a point mass, we have:

$$I_P^x = m_P\left(L_0^2 + (LC_{\alpha\theta})^2\right), \; I_P^y = m_P\left(LC_{\alpha\theta}C_\phi + L_0 S_\phi\right)^2, \; I_P^y = m_P\left(L_0 C_\phi - LS_\phi\right)^2. \qquad (42)$$

Likewise, using (34) and (35), $E_{Sl}^K$ takes the following form:

$$E_{Sl}^k = \tfrac{1}{2}m_{Sl}\left((R\dot{\theta}C_\phi)^2 + \left(\dot{\delta}C_\phi - (R+(\delta+\delta_0)S_\phi)\dot{\phi}\right)^2 + \left(\dot{\delta}S_\phi + (\delta+\delta_0)C_\phi\dot{\phi}\right)^2\right)$$
$$+ \tfrac{1}{2}I_{Sl}^x\dot{\phi}^2 + \tfrac{1}{2}I_{Sl}^y\left(\dot{\theta}C_\phi\right)^2 + \tfrac{1}{2}I_{Sl}^z\left(-\dot{\theta}S_\phi\right)^2, \qquad (43)$$

where considering the slider as a point mass, $I_{Sl}$ is derived as the following:

$$I_{Sl} = diag[I_{Sl}^x \; I_{Sl}^y \; I_{Sl}^z], \text{ with } I_{Sl}^x = m_{Sl}(\delta+\delta_0)^2, \; I_{Sl}^y = m_{Sl}\left((\delta+\delta_0)S_\phi\right)^2, \; I_{Sl}^z = m_{Sl}\left((\delta+\delta_0)C_\phi\right)^2. \qquad (44)$$

As the last step, substituting (37), (41), (43), and (20) in (14), and using moments of inertia that are presented in (40), (42), and (44), then, the Lagrange function can be written as:

$$\mathcal{L} = E_{Sp}^k + E_P^k + E_{Sl}^k - E^p$$
$$= \tfrac{1}{2}m_{Sp}\left((R\dot{\theta}C_\phi)^2 + (-R\dot{\phi})^2\right) + \tfrac{1}{2}\left(I_{Shl} + I_{Shf}^x\right)\dot{\phi}^2$$
$$+ \tfrac{1}{2}\left(I_{Shl} + I_{Shf}^x S_\phi^2\right)\left(\dot{\theta}C_\phi\right)^2 + \tfrac{1}{2}\left(I_{Shl} + I_{Shf}^x C_\phi^2\right)\left(-\dot{\theta}S_\phi\right)^2 \qquad (45)$$
$$+ \tfrac{1}{2}m_P\left(\left(R\dot{\theta}C_\phi + L(\dot{\alpha}-\dot{\theta})C_{\alpha\theta}\right)^2 + \left((L_0 S_\phi + LC_{\alpha\theta}C_\phi - R)\dot{\phi} - L(\dot{\alpha}-\dot{\theta})S_{\alpha\theta}S_\phi\right)^2\right.$$
$$\left. + \left(L(\dot{\alpha}-\dot{\theta})S_{\alpha\theta}C_\phi + (-L_0 C_\phi + LC_{\alpha\theta}S_\phi)\dot{\phi}\right)^2\right) + \tfrac{1}{2}m_P\left(L_0^2 + (LC_{\alpha\theta})^2\right)\dot{\phi}^2$$

$$+ \tfrac{1}{2} m_P \left( LC_{\alpha\theta}C_\phi + L_0 S_\phi \right)^2 \left( (\dot\theta - \dot\alpha) C_\phi \right)^2 + \tfrac{1}{2} m_P \left( L_0 C_\phi - L S_\phi \right)^2 \left( (\dot\alpha - \dot\theta) S_\phi \right)^2$$

$$+ \tfrac{1}{2} m_{Sl} \left( \left( R\dot\theta C_\phi \right)^2 + \left( \dot\delta C_\phi - (R + (\delta + \delta_0)) S_\phi ) \dot\phi \right)^2 + \left( \dot\delta S_\phi + (\delta + \delta_0) C_\phi \dot\phi \right)^2 \right)$$

$$+ \tfrac{1}{2} m_{Sl} (\delta + \delta_0)^2 \dot\phi^2 + \tfrac{1}{2} m_{Sl} \left( (\delta + \delta_0) S_\phi \right)^2 (\dot\theta C_\phi)^2 + \tfrac{1}{2} m_{Sl} \left( (\delta + \delta_0) C_\phi \right)^2 (-\dot\theta S_\phi)^2$$

$$- m_{Sl} g (\delta + \delta_0) S_\phi + m_P g \left( L C_{\alpha\theta} C_\phi + L_0 S_\phi \right).$$

Eventually, using (45) in (13), the equations of motion of Norma can be arranged to take the following form:

$$M(q)\ddot{q} + C(q,\dot{q})\dot{q} + G(q) = \mathcal{U}, \qquad (46)$$

where $M \in \mathbb{R}^{4\times 4}$ is the inertia matrix, $C \in \mathbb{R}^{4\times 4}$ is the Coriolis and centripetal matrix, and $G(q) \in \mathbb{R}^{4\times 1}$ is the vector of gravitational torques and forces. The elements of $\mathcal{U}$ take a zero value when there is a conservative motion in the associated coordinate. Hence, $\mathcal{U}_\theta = 0$, $\mathcal{U}_\alpha = \mathcal{T}$, $\mathcal{U}_\phi = 0$, $\mathcal{U}_\delta = \mathcal{F}$. The element-wise representation of (46) is:

$$\begin{bmatrix} M_{11} & M_{12} & M_{13} & M_{14} \\ M_{21} & M_{22} & M_{23} & M_{24} \\ M_{31} & M_{32} & M_{33} & M_{34} \\ M_{41} & M_{42} & M_{43} & M_{44} \end{bmatrix} \begin{bmatrix} \ddot\theta \\ \ddot\alpha \\ \ddot\phi \\ \ddot\delta \end{bmatrix} + \begin{bmatrix} C_{11} & C_{12} & C_{13} & C_{14} \\ C_{21} & C_{22} & C_{23} & C_{24} \\ C_{31} & C_{32} & C_{33} & C_{34} \\ C_{41} & C_{42} & C_{43} & C_{44} \end{bmatrix} \begin{bmatrix} \dot\theta \\ \dot\alpha \\ \dot\phi \\ \dot\delta \end{bmatrix} + \begin{bmatrix} G_1 \\ G_2 \\ G_3 \\ G_4 \end{bmatrix} = \begin{bmatrix} \mathcal{T} \\ 0 \\ 0 \\ \mathcal{F} \end{bmatrix}. \qquad (47)$$

In appendix A, the calculation steps for deriving the terms and elements in equations (47) have been elaborately presented.

## 4. Path Planning and Control of the Spherical Robot

In this section, the path tracking problem of the SR is investigated. First, a kinematics controller is designed that determines the desired values for the rolling angular velocity $\dot\theta_d$ and the slider displacement $\delta_d$. Then, a dynamics controller is introduced to govern $\mathcal{T}$ and $\mathcal{F}$ so the calculated desired values are achieved.

### A. Kinematics Control

To make the SR track a desired path, as the first step, a kinematics control scheme based on pure pursuit method [33] is proposed in this subsection. In Fig. 4, consider $\mathcal{P}$ as a two dimensional desired trajectory for the SR represented as $^{\mathcal{F}_W}[x_d(t), y_d(t)]^T$, that is assumed to be smooth, i.e. $\|^{\mathcal{F}_W}[\dot x_d, \dot y_d]^T\|$ is bounded for $t \geq 0$. It is desired that the location of $O_{Sp}$ in $\{\mathcal{F}_W\}$, $^{\mathcal{F}_W} r_{Sp} = ^{\mathcal{F}_W}[X_{Sp}(t), Y_{Sp}(t)]^T$, converges to $\mathcal{P}$ with a stable error.

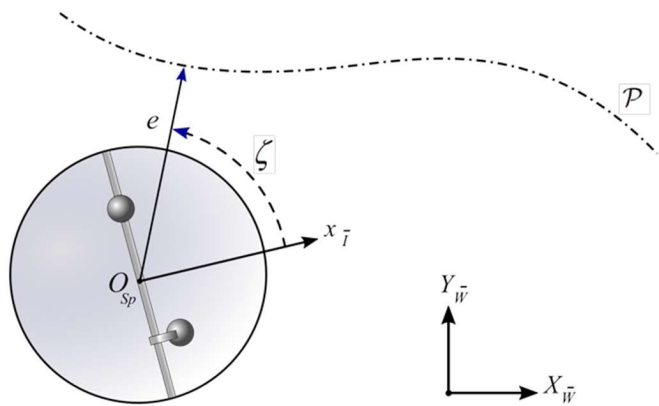

Fig. 4, The kinematics control for the spherical robot.

In Fig.4, let us define the tracking error vector as:

$$e = {}^{\mathcal{F}_W}P_{Sp} - {}^{\mathcal{F}_W}[x_d(t), y_d(t)]^T, \tag{48}$$

and the deviation angle between the SR's heading, ${}^{\mathcal{F}_W}x_l = [\cos(\psi), \sin(\psi)]^T$, and vector $e$ as:

$$\zeta = \angle(e, {}^{\mathcal{F}_W}x_l). \tag{49}$$

The objective of the kinematics control is to send the tracking error $e$ and also the deviation angle $\zeta$ to zero. To that end, we design the desired values of the slider displacement and the rolling angular velocity as follows:

$$\delta_d = k_1 \frac{\|e\|}{k_3 + \|e\|} \sin(\zeta), \text{ and} \tag{50}$$

$$\dot{\theta}_d = \frac{1}{R}\|[\dot{x}_d, \dot{y}_d]^T\| + k_2 \frac{\|e\|}{k_3 + \|e\|} \cos(\zeta). \tag{51}$$

where $k_1$, $k_2$, and $k_3$ are design parameters. In the above equations, $\|e\|/(k_3 + \|e\|)$ is used as a normalized error gain to prevent large control efforts at the vicinity of the target point. $\delta_d$ is selected to be directly proportional to $\sin(\zeta)$ as a measure of required tilting to compensate the deviation angle. In the design of $\dot{\theta}_d$, one can easily recognize that $\|[\dot{x}_d, \dot{y}_d]^T\|/R$ is the target's speed translated to the SR rolling velocity, $\dot{\theta}$. Therefore, this speed is chosen as the bias value for $\dot{\theta}_d$. To compensate the magnitude of $e$, the second term of $\dot{\theta}_d$ uses $\cos(\zeta)$ multiplied with the normalized error gain. $\cos(\zeta)$ is positive or negative when the SR is behind or ahead of the target, respectively. Now, that the SR's desired states are determined, the next step is to design a dynamic control scheme to track them.

*B. SR Dynamics Control*

In this step, a PID-based control scheme is presented for transforming $\dot{\theta}_d$ and $\delta_d$ to proper control actions in terms of $\mathcal{T}$ and $\mathcal{F}$ applied to the pendulum and slider, respectively. The control block diagram is depicted in Fig.5. As can be seen, two major control loops are designated for the tracking problem. The lower loop controls the slider's displacement. The applied force to the slider has two terms: (1) a force, equal to $gm_{Sl}\sin\phi$, against the gravity that is fed forward to prevent the slider from slipping when the SR is tilted, (2) a PID controller's output that is tuned to compensate the slider's displacement error. The upper loop controls the rolling angular velocity $\dot{\theta}$. To that end, first, an inner loop is designed to control the pendulum's angle, $\alpha - \theta$. Then, an outer loop is tuned to compensate the angular velocity error $\dot{\theta}_d - \dot{\theta}$ by governing the pendulum angle.

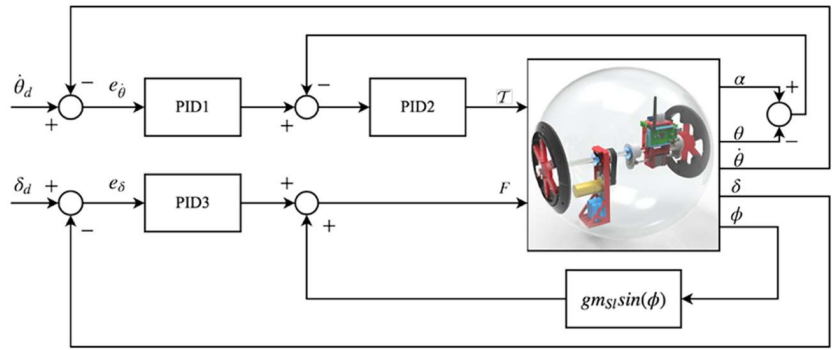

Fig. 5. The schematic of PID controller structure for the spherical robot.

# 5. SIMULATION RESULTS

In this section, the presented mathematical model of Norma is evaluated using the control scheme proposed in section 4. The SR parameters are chosen as follows: $R=0.2m$, $L=0.15m$, $L_0=0.10m$, $\delta_0=0.1m$, $m_{Sp}=3kg$, $m_{Shf}=m_P=m_{Sl}=0.5kg$, $g=9.81m/s^2$, $k_1=0.1$, $k_2=-0.1$, and $k_3=5$. Also, the PID controllers in Fig. 5 are tuned using values given in Table 1, which were obtained by trial-and-error.

Table 1: Numerical values of the PID controllers' parameters.

|      | $K_P$ | $K_I$ | $K_D$ |
|------|-------|-------|-------|
| **PID1** | 1  | 0 | 0  |
| **PID2** | 10 | 7 | 3  |
| **PID3** | 15 | 3 | 10 |

The simulations are carried out, where the desired trajectory $\mathcal{P}$ is chosen to be:

$$^{\mathcal{F}_W}[X_d,Y_d]=[2S_{0.01t},1.5S_{0.02t}] \text{ and } ^{\mathcal{F}_W}[X_{Sp}(0),Y_{Sp}(0)]=[0,0]. \tag{52}$$

The trajectory tracking performance of Norma in $XY$ plane is presented in Fig. 6. Fig. 7 shows the SR's closed-loop time response of the mathematical model. It can be observed that the SR can track the desired path successfully, and the tracking error converges to the vicinity of zero in a relatively short amount of time. The control actions governed by the PID controllers are given in Fig. 8. Finally, Figs. 9 and 10 illustrate the tilting angle and slider displacement during the analysis.

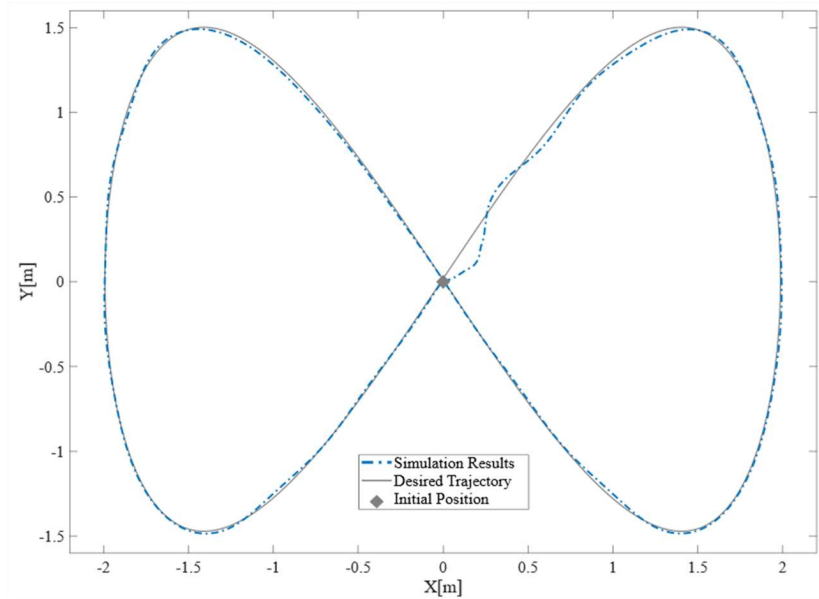

Fig. 6. Robot's desired and simulation trajectories in $XY$ plane.

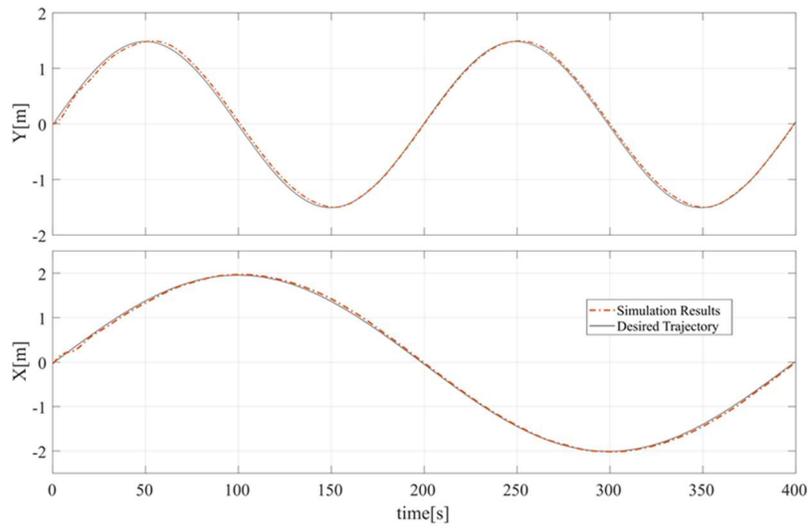

Fig. 7. SR trajectory in $X$ and $Y$ directions versus time.

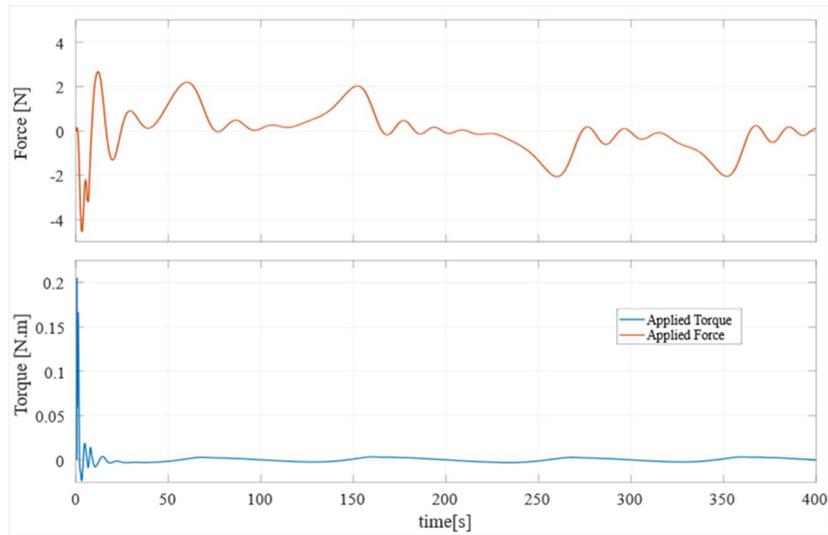

Fig. 8. Control actions governed by the controllers.

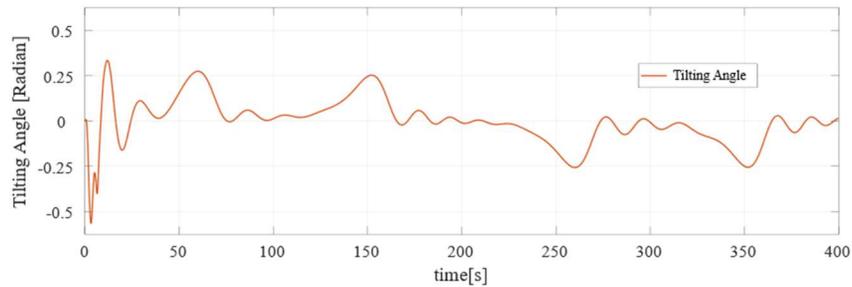

Fig. 9. Robot's Tilting angle.

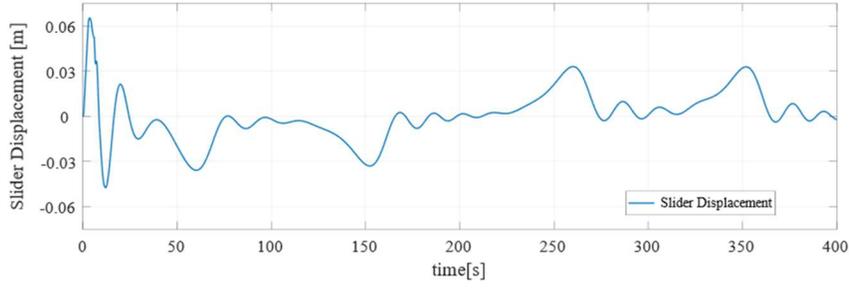
Fig. 10. Robot's Slider displacement.

## 6. CONCLUSION AND FUTURE WORKS

The dynamics of spherical robots are intrinsically complicated, posing challenges to derive their equations of motion. In this paper, we presented the conceptual design of Norma, a novel 2-DOF SR along with deriving its mathematical model. We also presented a kinematics and dynamics control scheme to track a desired trajectory by the robot and verified its performance through the simulation results.

The primary focus of this paper is to introduce the dynamics of the proposed SR. In another research paper [4], kinematics of Norma has been investigated on 3D terrains. One more step is to study the control problem of the derived dynamics in detail. Finally, incorporating the results of these studies, dynamics modeling, control, and 3D kinematics of Norma, the objective of this research project is to introduce a general model for the proposed SR.

## APPENDIX A

This appendix aims to show the calculation steps that have been carried out in order to derive the terms of the SR equations of motion presented in (47). Based on the Lagrange-Euler method, first, we calculate the $\partial \mathcal{L}/\partial \dot{q}_i$ term for each generalized coordinate $q$. Then, we collect the resulting expression in the following format:

$$\frac{\partial \mathcal{L}}{\partial \dot{q}_i} = \sum_{j=1}^{4} M_{ij}\dot{q}_j = M_{i1}\dot{\theta} + M_{i2}\dot{\alpha} + M_{i3}\dot{\phi} + M_{i4}\dot{\delta}. \tag{A.1}$$

For $q_1 = \theta$, the partial derivative of $\mathcal{L}$ with respect to $\dot{q}_1 = \dot{\theta}$ can be written in the form of:

$$\frac{\partial \mathcal{L}}{\partial \dot{\theta}} = M_{11}\dot{\theta} + M_{12}\dot{\alpha} + M_{13}\dot{\phi} + M_{14}\dot{\delta}, \tag{A.2}$$

where $M_{11}$ through $M_{14}$ can be calculated as:

$$M_{11} = L^2 m_P C_{\alpha\theta}^2 \left(1 + C_\phi^4\right) - 2LL_0 m_P C_\phi S_\phi^3 + 2L m_P C_{\alpha\theta} C_\phi \left(L_0 C_\phi^2 S_\phi - R\right) + L^2 m_P S_\phi^2 \left(S_{\alpha\theta}^2 + S_\phi^2\right)$$
$$+ C_\phi^2 \left(I_{Shl} + \left(m_P + m_{Sl} + m_{Sp}\right)R^2 + L^2 m_P S_{\alpha\theta}^2 + \left(I_{Shf}^x + 2\left(L_0^2 m_P + m_{Sl}(\delta_0 + \delta)^2\right)\right)S_\phi^2\right), \tag{A.3}$$

$$M_{12} = -m_P \left(L^2 C_{\alpha\theta}^2 \left(1 + C_\phi^4\right) - 2LL_0 C_\phi S_\phi^3 + LC_{\alpha\theta} C_\phi \left(2L_0 C_\phi^2 S_\phi - R\right) + L^2 S_\phi^2 \left(S_{\alpha\theta}^2 + S_\phi^2\right)\right.$$
$$\left. + C_\phi^2 \left(L^2 S_{\alpha\theta}^2 + 2L_0^2 S_\phi^2\right)\right), \tag{A.4}$$

$$M_{13} = L m_P S_{\alpha\theta}\left(L_0 - RS_\phi\right), \tag{A.5}$$

$$M_{14} = 0. \tag{A.6}$$

For $q_2 = \alpha$, $\partial \mathcal{L}/\partial \dot{\alpha}$ will be:

$$\frac{\partial \mathcal{L}}{\partial \dot{\alpha}} = M_{21}\dot{\theta} + M_{22}\dot{\alpha} + M_{23}\dot{\phi} + M_{24}\dot{\delta}, \tag{A.7}$$

where terms $M_{21}$ through $M_{24}$ are as follows:

$$M_{21} = M_{12}, \tag{A.8}$$

$$M_{22} = m_P\left(L^2 C_{\alpha\theta}^2 + L^2 C_\phi^2 S_{\alpha\theta}^2 + S_\phi^2\left((L_0 C_\phi - LS_\phi)^2 + L^2 S_{\alpha\theta}^2\right) + C_\phi^2\left(LC_{\alpha\theta}C_\phi + L_0 S_\phi\right)^2\right), \tag{A.9}$$

$$M_{23} = -Lm_P S_{\alpha\theta}(L_0 - RS_\phi), \tag{A.10}$$

$$M_{24} = 0. \tag{A.11}$$

For $q_3 = \phi$ we have:

$$\frac{\partial \mathcal{L}}{\partial \dot{\phi}} = M_{31}\dot{\theta} + M_{32}\dot{\alpha} + M_{33}\dot{\phi} + M_{34}\dot{\delta}, \tag{A.12}$$

where $M_{31}$ through $M_{34}$ are as the following:

$$M_{31} = M_{13}, \tag{A.13}$$

$$M_{32} = M_{23}, \tag{A.14}$$

$$M_{33} = I_{Shl} + I_{Shf}^x + 2m_{Sl}(\delta_0 + \delta)^2 + (m_P + m_{Sl} + m_{Sp})R^2 + m_P\left(L^2(1 + C_{2\alpha\theta}) + 2L_0^2 - 2RLC_{\alpha\theta}C_\phi\right) \\ + 2R(m_{Sl}(\delta_0 + \delta) - L_0 m_P)S_\phi, \tag{A.15}$$

$$M_{34} = -m_{Sl} R C_\phi. \tag{A.16}$$

Finally, for $q_4 = \delta$, $\partial \mathcal{L}/\partial \dot{\delta}$ is written as:

$$\frac{\partial \mathcal{L}}{\partial \dot{\delta}} = M_{41}\dot{\theta} + M_{42}\dot{\alpha} + M_{43}\dot{\phi} + M_{44}\dot{\delta}, \tag{A.17}$$

with

$$M_{41} = M_{14}, \tag{A.18}$$

$$M_{42} = M_{24}, \tag{A.19}$$

$$M_{43} = M_{34}, \text{ and} \tag{A.20}$$

$$M_{44} = m_{Sl}. \tag{A.21}$$

The next step is to derive the terms of $C(q,\dot{q})$ and $G(q)$ to form the equation (46). It can be shown that there is not a unique arrangement for the elements in $C$. For controller design applications, it is preferred to form $C$ such that the matrix

$$N(q,\dot{q}) = \dot{M}(q,\dot{q}) - 2C(q,\dot{q}), \tag{A.22}$$

be a skew-symmetric, i.e. $N^T = -N$. To that end, the $C$ elements are determined as:

$$C_{ij} = \sum_{k=1}^{4} c_{kji}\dot{q}_k, \tag{A.23}$$

where,

$$c_{kji} = \frac{1}{2}\left(\frac{\partial M_{ij}}{\partial q_k} + \frac{\partial M_{ik}}{\partial q_j} - \frac{\partial M_{kj}}{\partial q_i}\right). \tag{A.24}$$

$c_{kji}$ are the so-called Christoffel [34] symbols of the first kind. Therefore, using (A.23) and (A.24), for $q_1 = \theta$, terms of $C_{11}$ through $C_{14}$ are calculated as:

$$C_{11} = C_\phi \left( Lm_P (\dot\alpha - \dot\theta) S_{\alpha\theta} \left( R - C_\phi^2 (LC_{\alpha\theta} C_\phi + L_0 S_\phi) \right) m_{Sl} (\delta_0 + \delta) \dot\delta S_\phi S_{2\phi} \right) + \tfrac{1}{8} \dot\phi \big( LL_0 m_P C_{(\alpha\theta)/2}^2 C_{4\phi}$$
$$+ 8Lm_P \left( -L_0 C_{2\phi} S_{(\alpha\theta)/2}^2 + RC_{\alpha\theta} S_\phi \right) - 2 S_{2\phi} \left( 2I_{Shl} + 2(m_P + m_{Sl} + m_{Sp}) R^2 + L^2 m_P (C_{2\alpha\theta} - 1) \right) \quad (A.25)$$
$$+ S_{4\phi} \left( 2I_{Shf}^x + 4L_0^2 m_P + 4 m_{Sl} (\delta_0 + \delta)^2 - L^2 m_P (3 + C_{2\alpha\theta}) \right),$$

$$C_{12} = Lm_P C_\phi (\dot\alpha - \dot\theta) S_{\alpha\theta} \left( LC_{\alpha\theta} C_\phi^3 + L_0 C_\phi^2 S_\phi - R \right) + \tfrac{1}{4} m_P \dot\phi \big( 8L^2 C_{\alpha\theta}^2 C_\phi^3 S_\phi$$
$$- 2LC_{\alpha\theta} \left( L_0 (C_{2\phi} + C_{4\phi}) + RS_\phi \right) \quad (A.26)$$
$$+ 4LS_\phi \left( L_0 S_{3\phi} - LC_\phi \right) + S_{4\phi} \left( L^2 - 2L_0^2 \right) \big),$$

$$C_{13} = \tfrac{1}{8} \big( 8LL_0 m_P C_{(\alpha\theta)/2}^2 C_{4\phi} (\dot\theta - \dot\alpha) + 8LL_0 m_P C_{2\phi} (\dot\alpha - \dot\theta) S_{(\alpha\theta)/2}^2$$
$$- 8L^2 m_P \dot\phi S_{2\alpha\theta} - 4Lm_P RC_{2\alpha\theta} (\dot\alpha - 2\dot\theta) S_\phi$$
$$+ 2 \big( (L^2 m_P (1 - C_{2\alpha\theta}) - 2I_{Shl} - 2(m_P + m_{Sl} + m_{Sp}) R^2 ) \dot\theta - 2L^2 m_P \dot\alpha S_{\alpha\theta}^2 \big) S_{2\phi} + \big( m_P \dot\alpha (L^2 (3 + C_{2\alpha\theta}) - 4L_0^2) \quad (A.27)$$
$$+ \big( 2I_{Shl} - 3L^2 m_P + 4L_0^2 m_P + 4 m_{Sl} (\delta_0 + \delta)^2 - L^2 m_P C_{2\alpha\theta} \big) \dot\theta \big) S_{4\phi} \big),$$

$$C_{14} = 2 m_{Sl} \dot\theta (\delta_0 + \delta) C_\phi^2 S_\phi^2. \quad (A.28)$$

Using the same method, for $q_2 = \alpha$ we can calculate $C_{21}$ through $C_{24}$ as the following:

$$C_{21} = Lm_P C_\phi^3 (\dot\alpha - \dot\theta) S_{\alpha\theta} \left( LC_{\alpha\theta} C_\phi + L_0 S \right) + \tfrac{1}{4} m_P \dot\phi \big( 8L^2 C_{\alpha\theta}^2 C_\phi^3 S_\phi - 2LC_{\alpha\theta} \left( L_0 (C_{2\phi} + C_{4\phi}) + RS_\phi \right)$$
$$+ 4LS_\phi \left( L_0 S_{3\phi} - LC_\phi \right) + \left( L^2 - 2L_0^2 \right) S_{4\phi} \big), \quad (A.29)$$

$$C_{22} = -Lm_P C_\phi^3 (\dot\alpha - \dot\theta) S_{\alpha\theta} \left( LC_{\alpha\theta} C_\phi + L_0 S_\phi \right) - \tfrac{1}{4} m_P \dot\phi \big( -4LC_\phi \left( L_0 C_{\alpha\theta} C_{3\phi} + LS_\phi - 2LC_{\alpha\theta}^2 C_\phi^2 S_\phi \right)$$
$$+ 4LL_0 S_\phi S_{3\phi} + \left( L^2 - 2L_0^2 \right) S_{4\phi} \big), \quad (A.30)$$

$$C_{23} = \tfrac{1}{8} m_P \big( 8LL_0 C_{\alpha\theta}^2 C_{4\phi} (\dot\alpha - \dot\theta) + 8LL_0 C_{2\phi} (\dot\theta - \dot\alpha) S_{\alpha\theta}^2 + 8L^2 \dot\phi S_{2\alpha\theta} - 4LRC_{\alpha\theta} \dot\theta S_\phi$$
$$+ 4L^2 (\dot\alpha - \dot\theta) S_{\alpha\theta}^2 S_{2\phi} \quad (A.31)$$
$$- \left( L^2 (3 + C_{2\alpha\theta}) - 4L_0^2 \right) (\dot\alpha - \dot\theta) S_{4\phi} \big),$$

$$C_{24} = 0. \quad (A.32)$$

Next, for $q_3 = \phi$, $C_{31}$ trough $C_{34}$ are as follows:

$$C_{31} = Lm_P (2LC_{\alpha\theta} - RC_\phi) \dot\phi S_{\alpha\theta} + \tfrac{1}{4} m_P \dot\alpha \big( 2L(-4LC_{\alpha\theta}^2 C_\phi^3 S_\phi + C_{\alpha\theta} \left( L_0 (2 + C_{2\phi} + C_{4\phi}) - RS_\phi \right) + LS_{2\phi}$$
$$- 2L_0 S_\phi S_{3\phi}) - (L^2 - 2L_0^2) S_{4\phi} \big) + \tfrac{1}{8} \dot\theta \big( -8LL_0 m_P \left( C_{\alpha\theta} + C_{\alpha\theta/2}^2 C_{4\phi} \right) + 8LL_0 m_P C_{2\phi} S_{\alpha\theta/2}^2$$
$$+ 2 \big( 2I_{Shl} + L^2 m_P (C_{2\alpha\theta} - 1) + 2R^2 (m_P + m_{Sl} + m_{Sp}) \big) S_{2\phi} - \big( 2I_{Shf}^x - 3L^2 m_P + 4L_0^2 m_P + 4 m_{Sl} (\delta + \delta_0)^2 \quad (A.33)$$
$$- L^2 m_P C_{2\alpha\theta} \big) S_{4\phi} \big),$$

$$C_{32} = 1/8 m_P \big( 8LL_0 \left( C_{\alpha\theta/2}^2 C_{4\phi} (\dot\theta - \dot\alpha) + 8LL_0 C_{2\phi} (\dot\alpha - \dot\theta) S_{\alpha\theta/2}^2 + 8LR\dot\phi C_\phi S_{\alpha\theta}$$
$$+ 8LC_{\alpha\theta} \left( L_0 \dot\theta - L_0 \dot\alpha - 2L\dot\phi S_{\alpha\theta} \right) \quad (A.34)$$
$$4LRC_{\alpha\theta} (2\dot\alpha - \dot\theta) S_\phi + 4L^2 (\dot\theta - \dot\alpha) S_{\alpha\theta}^2 S_{2\phi} + \left( L^2 (C_{2\alpha\theta} + 3) - 4L_0^2 \right) (\dot\alpha - \dot\theta) S_{4\phi} \big),$$

$$C_{33} = Lm_P\left(-2LC_{\alpha\theta} + RC_\phi\right)\left(\dot\alpha - \dot\theta\right)S_{\alpha\theta} + m_{Sl}\dot\delta\left(2(\delta_0+\delta) + RsS_\phi\right)$$
$$+ R\dot\phi\left(\left(m_{Sl}(\delta_0+\delta) - L_0m_P\right)C_\phi + Lm_P C_{\alpha\theta}S_\phi\right), \tag{A.35}$$

$$C_{34} = m_{Sl}\dot\phi\left(2(\delta_0+\delta) + RS_\phi\right). \tag{A.36}$$

Finally, for $q_4 = \delta$, $C_{41}$ trough $C_{44}$ are:

$$C_{41} = -2m_{Sl}(\delta_0+\delta)C_\phi^2 \dot\theta S_\phi^2, \tag{A.37}$$

$$C_{42} = 0, \tag{A.38}$$

$$C_{43} = -2m_{Sl}(\delta_0+\delta)\dot\phi, \tag{A.39}$$

$$C_{44} = 0. \tag{A.40}$$

The last step is to calculate $G(q)$, using the following equation,

$$G_i(q) = \frac{\partial E^p}{\partial q_i}. \tag{A.41}$$

for $q_1 = \theta$ through $q_4 = \delta$ we have:

$$G_1 = -gLm_P C_\phi S_{\alpha\theta}, \tag{A.42}$$

$$G_2 = gLm_P C_\phi S_{\alpha\theta}, \tag{A.43}$$

$$G_3 = g\left(\left(m_{Sl}(\delta_0+\delta) - L_0m_P\right)C_\phi + Lm_P C_{\alpha\theta}S_\phi\right), \tag{A.44}$$

$$G_4 = gm_{Sl}S_\phi. \tag{A.45}$$

By substituting calculated terms of the matrix $M(q)$, $C(q,\dot q)$, and $G(q)$ in (47), the dynamics model for the presented spherical robot is complete.

FUNDING, DATA AVAILABILITY, AND CONFLICTS OF INTEREST STATEMENT

This research was funded by the college of engineering of Lamar University.

The data used to support the findings of this study are available from the corresponding author upon request.

The authors declare that they have no conflicts of interest.